\documentclass[11pt]{article}

\usepackage[utf8]{inputenc}
\usepackage{amsmath}
\usepackage{amssymb}
\usepackage{graphicx}
\usepackage[numbers,sort&compress]{natbib}
\usepackage{geometry}
\usepackage{booktabs}
\usepackage{longtable}
\usepackage{array}
\usepackage{caption}
\usepackage{hyperref}
\usepackage{algorithm}
\usepackage{algpseudocode}

\hypersetup{
    colorlinks=true,
    linkcolor=blue,
    filecolor=magenta,
    urlcolor=cyan,
    citecolor=blue
}
\usepackage{parskip}
\usepackage{enumitem}
\usepackage{rotating}

\title{Scalable and Interpretable Contextual Bandits: A Literature Review and Retail Offer Prototype}
\author{
    Nikola Tanković \\
    Juraj Dobrila University of Pula, Faculty of Informatics \\
    \texttt{ntankov@unipu.hr}
    \and
    Robert Šajina \\
    Juraj Dobrila University of Pula, Faculty of Informatics \\
    \texttt{rosaj@unipu.hr}
}
\date{May 20, 2025}

\begin{document}
\maketitle
\begin{abstract}
This paper presents a concise review of Contextual Multi-Armed Bandit (CMAB) methods and introduces an experimental framework for scalable, interpretable offer selection, addressing the challenge of fast-changing offers. The approach models context at the product category level, allowing offers to span multiple categories and enabling knowledge transfer across similar offers. This improves learning efficiency and generalization in dynamic environments. The framework extends standard CMAB methodology to support multi-category contexts, and achieves scalability through efficient feature engineering and modular design. Advanced features such as MPG (Member Purchase Gap) and MF (Matrix Factorization) capture nuanced user-offer interactions, with implementation in Python for practical deployment.

A key contribution is interpretability at scale: logistic regression models yield transparent weight vectors, accessible via a large language model (LLM) interface for real-time, user-level tracking and explanation of evolving preferences. This enables the generation of detailed member profiles and identification of behavioral patterns, supporting personalized offer optimization and enhancing trust in automated decisions. By situating our prototype alongside established paradigms like Generalized Linear Models and Thompson Sampling, we demonstrate its value for both research and real-world CMAB applications.
\end{abstract} 

\section{Introduction}

Contextual Multi-Armed Bandits (CMABs) provide a powerful framework for sequential decision-making under uncertainty, where an agent learns to select actions (arms) based on observed contextual information to maximize cumulative rewards over time \cite{Chen2024Survey, Slivkins2019IntroductionMAB}. This paradigm addresses the fundamental exploration-exploitation dilemma \cite{Slivkins2019IntroductionMAB, Auer2002FiniteTime}, requiring a balance between trying new actions to gather information and choosing actions known to yield high rewards. The ability of CMABs to leverage side information for personalized decisions has led to their widespread adoption in diverse applications, including personalized recommendation systems for news \cite{Li2010Contextual}, movies \cite{Chen2024ContextualBanditsMovieRec}, products \cite{Xia2024RetailSynth}, online advertising \cite{Slivkins2019IntroductionMAB, Badanidiyuru2014Resourceful}, dynamic pricing \cite{Li2010Contextual, Badanidiyuru2014Resourceful, Javanmard2021OnlineMarkdown}, and medical treatment optimization \cite{Li2010Contextual, Slivkins2019IntroductionMAB}. The field has seen a significant evolution, from foundational statistical models to sophisticated deep learning approaches \cite{Chen2024Survey, Slivkins2019IntroductionMAB}, reflecting a broader trend in machine learning towards harnessing large datasets and computational power for more nuanced and effective automated decision-making. This progression mirrors the increasing complexity of real-world problems, where simple linear relationships often fall short, necessitating models that can capture intricate patterns in high-dimensional contextual data. Several comprehensive surveys and book chapters offer detailed accounts of CMAB algorithms, their theoretical underpinnings, and practical applications \cite{Slivkins2019IntroductionMAB, LattimoreSzepesvari2020BanditAlgorithms, Chen2024Survey}.

\section{Literature Review}
The CMAB problem has been tackled by several families of algorithms, each with distinct mechanisms for balancing exploration and exploitation and for modeling the relationship between contexts, actions, and rewards.

\subsection{Upper Confidence Bound (UCB) Strategies}
UCB-based algorithms operationalize the principle of "optimism in the face of uncertainty" (OFU) \cite{Auer2002Associative, AbbasiYadkori2011Improved, Han2024UCBRegret}. At each decision point, these algorithms estimate the expected reward for each arm and construct an upper confidence bound (UCB) around this estimate. The arm with the highest UCB is selected, thereby encouraging exploration of arms whose rewards are uncertain but potentially high \cite{Auer2002FiniteTime, Li2010Contextual, ContextualBanditsLibrary}.

A seminal algorithm in this category is LinUCB, introduced by Li et al. \cite{Li2010Contextual} and further analyzed by Chu et al. \cite{Chu2011Contextual, AbbasiYadkori2011Improved}. LinUCB models the expected reward $E[r_{t,a} | x_{t,a}]$ for arm $a$ at time $t$ as a linear function of its $d$-dimensional feature vector, using either a disjoint ($x_{t,a}^T \theta_a^*$) or shared ($x_{t,a}^T \theta^*$) parameterization \cite{Li2010Contextual, Chu2011Contextual}. The algorithm applies ridge regression to estimate the unknown parameter(s) and adds an exploration bonus proportional to the standard deviation, scaled by a hyperparameter $\alpha$ \cite{Li2010Contextual}. LinUCB achieved notable improvements in click-through rates for news recommendation \cite{Li2010Contextual}, and its variants are used in commercial systems \cite{ContextualBanditsLibrary}. Theoretical analyses provide regret bounds of order $\tilde{O}(d\sqrt{T})$ \cite{Chu2011Contextual, AbbasiYadkori2011Improved, SimchiLeviXu2021Bypassing}. Abbasi-Yadkori et al. further refined these results with algorithms like OFUL, which offer tighter confidence bounds and improved regret guarantees \cite{AbbasiYadkori2011Improved, LattimoreSzepesvari2020BanditAlgorithms}.

The inherent simplicity and provable guarantees of LinUCB under linearity assumptions made it a foundational model. However, the constraint of linearity is often too restrictive for real-world reward landscapes. This limitation spurred research into UCB methods for more expressive models. Generalized Linear Models (GLMs) were a natural extension, with algorithms like LogisticUCB \cite{ContextualBanditsLibrary, Filippi2010Parametric} designed for binary outcomes (e.g., clicks), using a logistic link function. Filippi et al. provided early work on UCB for GLMs \cite{Filippi2010Parametric}. More recently, neural network-based UCB approaches, such as NeuralUCB \cite{Zhou2020NeuralUCB} and its robust variant R-NeuralUCB \cite{Zhou2020NeuralUCB}, have been developed to capture complex non-linear context-reward relationships by leveraging the representation power of deep learning. Constructing theoretically valid confidence bounds for these highly non-linear models remains an active research area, with techniques like using network gradients for exploration being explored \cite{Zhou2020NeuralUCB}. Other UCB variants address specific challenges: D-LinUCB adapts to non-stationary environments via discounted regression \cite{Garcelon2020DLinUCB}, Hellinger-UCB shows promise for cold-start scenarios by using Hellinger distance in UCB construction \cite{Jiang2024HellingerUCB}. This trajectory from linear to increasingly complex models underscores a persistent theme: the quest for algorithms that are both powerful enough for real-world data and amenable to theoretical performance guarantees.

\subsection{Epsilon-Greedy ($\epsilon$-Greedy) Methods}
The $\epsilon$-Greedy strategy offers a simpler alternative for balancing exploration and exploitation. With probability $1-\epsilon$, the agent selects the action with the currently highest estimated reward (exploitation), and with probability $\epsilon$, it selects an action uniformly at random from all available actions (exploration) \cite{Slivkins2019IntroductionMAB, SuttonBarto1998RL}. Its ease of implementation has made it a popular choice in practice \cite{SuttonBarto1998RL, LangfordZhang2008EpochGreedy}, and it can be paired with any underlying reward prediction model, such as logistic regression for predicting click probabilities \cite{SuttonBarto1998RL}.

Several variants aim to improve upon the basic $\epsilon$-Greedy mechanism. A common approach is to use a decaying $\epsilon$, where the probability of exploration decreases over time, leading to more exploitation as the agent gains confidence in its reward estimates \cite{Slivkins2019IntroductionMAB}. Adaptive $\epsilon$ strategies adjust $\epsilon$ based on measures of learning progress or the observed differences in action values \cite{Slivkins2019IntroductionMAB}; the \emph{contextual-bandits} Python library, for instance, implements \emph{AdaptiveGreedy} alongside the standard \emph{EpsilonGreedy} \cite{ContextualBanditsLibrary}. While fundamentally simple, the theoretical understanding of $\epsilon$-Greedy in complex reinforcement learning settings, particularly those involving function approximation, is an area of ongoing research \cite{LangfordZhang2008EpochGreedy, Vermeulen2020EpsilonGreedyRL}. The enduring appeal and continued adaptation of $\epsilon$-Greedy, despite its known theoretical limitations in certain scenarios compared to UCB or Thompson Sampling, underscore a practical demand for computationally inexpensive and straightforward bandit solutions. When the complexity of implementing and tuning more sophisticated exploration mechanisms is a barrier, $\epsilon$-Greedy provides a basic yet often effective approach. The development of adaptive epsilon strategies attempts to imbue this simple heuristic with more intelligence, reducing the gap to more advanced methods \cite{Tokic2010AdaptiveEpsilon}.

\subsection{Posterior Sampling}
Thompson Sampling (TS), also known as posterior sampling or probability matching, is a Bayesian approach to the exploration-exploitation trade-off \cite{Russo2018TutorialTS, ChapelleLi2011EmpiricalTS, AgrawalGoyal2013Thompson}. It maintains a posterior distribution over the parameters of the reward model. In each round, the algorithm samples a parameter vector from this posterior and then selects the action that is optimal according to these sampled parameters \cite{Russo2018TutorialTS, ChapelleLi2011EmpiricalTS}. This inherently random process, driven by the uncertainty in the posterior, facilitates exploration.

TS has garnered significant attention due to its strong empirical performance, often matching or exceeding UCB-based methods in practical applications \cite{Russo2018TutorialTS, ChapelleLi2011EmpiricalTS, AgrawalGoyal2012ThompsonAnalysis}, as highlighted in the influential empirical evaluation by Chapelle and Li \cite{ChapelleLi2011EmpiricalTS}. Theoretical understanding, particularly for contextual settings, initially lagged behind its empirical success. Agrawal and Goyal provided a key breakthrough with theoretical guarantees for TS in linear contextual bandits, proving a regret bound of $\tilde{O}(d^{3/2}\sqrt{T})$ using Gaussian priors and likelihoods \cite{AgrawalGoyal2013Thompson, ParkFaradonbeh2024ThompsonPartially}. This work was crucial in establishing the theoretical soundness of TS in this important setting.

Thompson Sampling is readily adaptable to various reward models. For logistic rewards, common in click-through rate prediction, PG-TS (Pólya-Gamma Thompson Sampling) leverages Pólya-Gamma data augmentation for efficient Gibbs sampling to approximate the posterior \cite{Dumitrascu2018PGTS}. The \emph{contextual-bandits} library also includes an implementation of \emph{LogisticTS} \cite{ContextualBanditsLibrary}. Further theoretical advancements include the Generalized Thompson Sampling framework by Li (2013), which situates TS within an expert-learning context, offering regret bounds for general contextual bandits and a way to quantify the impact of the prior distribution \cite{Li2013GeneralizedTS}. Specialized TS algorithms have also been developed, such as TS-Conf, which aims to mitigate herding effects in user feedback within recommendation systems \cite{Zhang2024TSConf}, and DISCO, which integrates TS with integer programming to handle budget constraints in personalized discount allocation \cite{ZhangHowson2024DISCO}. The strength of Thompson Sampling often lies in its natural and effective way of encoding uncertainty into the exploration process. By sampling from the posterior, it automatically explores more uncertain actions (those with wider posteriors) without explicit exploration bonuses, leading to robust performance across a wide range of problems.

\subsection{Logistic and Generalized Linear Bandits (GLMs)}
While linear bandit models assume a direct linear relationship between features and expected rewards, Generalized Linear Models (GLMs) extend this to accommodate non-linear relationships and specific reward distributions, such as binary outcomes for clicks or count data for purchases, through the use of a link function \cite{Slivkins2019IntroductionMAB, ZhangSugiyama2023OnlineMultinomial, Filippi2010Parametric}.

Logistic bandits are a prominent type of GLM where the probability of a binary reward (e.g., click=1, no-click=0) is modeled using the logistic (sigmoid) function: $P(r=1 | x) = \sigma(x^T \theta^*) = 1 / (1 + \exp(-x^T \theta^*))$ \cite{ZhangSugiyama2023OnlineMultinomial, Filippi2010Parametric}. This model is particularly well-suited for applications like click-through rate (CTR) optimization. Algorithmic approaches for logistic bandits often build upon UCB principles (e.g., LogisticUCB \cite{ContextualBanditsLibrary}, and the work of Filippi et al. (2010) \cite{Filippi2010Parametric} on UCB for GLMs) or Thompson Sampling (e.g., LogisticTS \cite{ContextualBanditsLibrary}, PG-TS \cite{Dumitrascu2018PGTS}).

A key development in this area is the reduction of contextual bandit problems to online regression. Foster and Krishnamurthy demonstrated an efficient reduction to online regression with the logarithmic loss, achieving first-order regret guarantees, which can be significantly better than worst-case bounds on "easy" instances \cite{FosterKrishnamurthy2021Efficient, SimchiLeviXu2021Bypassing}. This work highlighted that the choice of loss function for the underlying online regressor (e.g., log loss versus square loss) is critical for attaining certain types of regret bounds, revealing deep connections between bandit theory and online learning. More recently, Zhang and Sugiyama (2023) proposed OFUL-MLogB for multinomial logistic bandits (handling more than two outcomes), achieving improved regret bounds and constant per-round computation cost, notably removing the dependency on a problem-specific constant $\kappa$ that could be exponentially large in prior work \cite{ZhangSugiyama2023OnlineMultinomial}. This shift towards GLMs, and logistic bandits in particular, reflects a practical necessity to model discrete, bounded, or probabilistic rewards more faithfully than is possible with simple linear models.

L2 regularization is commonly employed within the underlying regression components of many contextual bandit algorithms, including LinUCB (which often uses ridge regression) \cite{Li2010Contextual, ContextualBanditsLibrary} and logistic regression models \cite{Dumitrascu2018PGTS, ContextualBanditsLibrary, SuttonBarto1998RL}. Regularization helps prevent overfitting, especially in high-dimensional feature spaces, and improves the stability and generalization of the learned models. The impact of regularization in online learning and bandit settings is a well-studied area \cite{SuttonBarto1998RL, Hazan2016IntroductionOnlineConvex, Zinkevich2003OnlineConvex, Xiao2010Dual, Bastani2020OnlineDecision, FosterRakhlin2020BeyondUCB, SimchiLeviXu2021Bypassing}, with L2 regularization being a standard technique to control model complexity and ensure smoother parameter estimates. Official Python implementation \emph{contextual-bandits} \cite{ContextualBanditsLibrary} explicitly details L2 regularization ($\lambda$) in the LinUCB update equations, and \cite{SuttonBarto1998RL} discusses its use in Online Logistic Regression for Thompson Sampling.

\begin{sidewaystable}
\centering
\caption{Summary of Key Contextual Bandit Algorithm Families}
\label{tab:algo_summary}
\begin{tabular}{p{2.5cm}p{3cm}p{4.5cm}p{2.5cm}p{3.5cm}}
\toprule
\textbf{Approach} & \textbf{Algorithm} & \textbf{Core Principle} & \textbf{Reward Model} & \textbf{Key Reference(s)} \\
\midrule
UCB & LinUCB & Optimism in Face of Uncertainty (OFU), linear reward estimation + confidence bound & Linear & Li et al. (2010) \cite{Li2010Contextual}, Chu et al. (2011) \cite{Chu2011Contextual} \\
& LogisticUCB & OFU, logistic reward estimation + confidence bound & Logistic (GLM) & Filippi et al. (2010) \cite{Filippi2010Parametric}, contextual-bandits lib \cite{ContextualBanditsLibrary} \\
& NeuralUCB & OFU, neural network reward estimation + gradient-based exploration bonus & Non-linear (NN) & Zhou et al. (2020) \cite{Zhou2020NeuralUCB} \\
\midrule
Epsilon-Greedy & $\epsilon$-Greedy & Probabilistic exploration (random action with prob $\epsilon$), greedy exploitation & Agnostic (uses any reward predictor) & Sutton \& Barto (1998) (general RL) \cite{SuttonBarto1998RL}, Langford \& Zhang (2008) (contextual) \cite{LangfordZhang2008EpochGreedy} \\
\midrule
Thompson Sampling & LinearTS & Posterior sampling, choose action optimal for sampled linear parameters & Linear (often Gaussian likelihood) & Agrawal \& Goyal (2013) \cite{AgrawalGoyal2013Thompson} \\
& LogisticTS & Posterior sampling, choose action optimal for sampled logistic parameters & Logistic (GLM) & Chapelle \& Li (2011) \cite{ChapelleLi2011EmpiricalTS}, PG-TS \cite{Dumitrascu2018PGTS} \\
\midrule
GLM-based & OFUL-MLogB & OFU for multinomial logistic regression, efficient online estimation & Multinomial Logistic (GLM) & Zhang \& Sugiyama (2023) \cite{ZhangSugiyama2023OnlineMultinomial} \\
& FastCB & Reduction to online regression with log loss for first-order regret & Agnostic (via oracle) & Foster \& Krishnamurthy (2021) \cite{FosterKrishnamurthy2021Efficient} \\
\bottomrule
\end{tabular}
\end{sidewaystable}

\section{The Role of Context: Feature Engineering and Representation}
The "context" in CMABs is operationalized through feature vectors, and the quality, relevance, and representation of these features are paramount to the algorithm's success \cite{Chen2024Survey, Lin2018AdaptiveFeature, Hao2024FeatureSelectionCB}. Effective feature engineering can significantly enhance a bandit's ability to discern optimal actions in varying situations, while poor or irrelevant features can degrade performance or lead to overfitting \cite{Hao2024FeatureSelectionCB}.

Common categories of features include user-specific attributes like demographics, historical interactions (e.g., purchase history, browsing logs), and derived preferences or loyalty scores \cite{Chen2024Survey, Bastani2020OnlineDecision}. Item or arm features describe the characteristics of the available actions, such as product details (category, price, brand), offer specifics (discount value, type of promotion), or content attributes (genre, keywords) \cite{Chen2024ContextualBanditsMovieRec, ContextualBanditsLibrary, ZhangHowson2024DISCO}. Interaction features, which capture the interplay between user and item characteristics (e.g., a user's affinity for a particular product category), are often critical for personalization and can be explicitly engineered or implicitly learned by more complex models \cite{Chen2024Survey, Cheng2016WideDeep, Tang2014EnsembleContextual}. Temporal and session-based features, such as time of day, day of week, seasonality, visit recency, session duration, and real-time in-session behavior (e.g., items currently in a shopping cart), are crucial for capturing the dynamic and often non-stationary nature of user behavior and environmental conditions \cite{Chen2024Survey, Bastani2020OnlineDecision, ZHAO2024107558, Kooti2016Portrait, Park2024EMKFBandit, Wang2024LinUCBHybrid}. For instance, Kooti et al. (2016) demonstrated the high predictive power of temporal features for forecasting the timing of a user's next purchase \cite{Kooti2016Portrait}.

Representing these diverse features effectively is a key challenge. Basic techniques include one-hot encoding for discrete variables and normalization or standardization for continuous ones \cite{Chen2024Survey, ContextualBanditsLibrary}. For high-cardinality categorical features (e.g., user IDs, product IDs), embeddings are widely used to learn dense, lower-dimensional vector representations that can capture semantic relationships \cite{Chen2024Survey}. Deep learning models often incorporate embedding layers as an integral part of their architecture \cite{Cheng2016WideDeep, Zhou2020NeuralUCB}.

More advanced feature engineering strategies involve adaptive and learned representations. Lin et al. proposed ABaCoDE, a method that employs online clustering and autoencoders for adaptive feature extraction. It pre-trains on unlabeled historical context data and then selects and adapts encoders online, showing particular benefits in non-stationary environments \cite{Lin2018AdaptiveFeature}. Deep learning architectures like Convolutional Neural Networks (CNNs) and Recurrent Neural Networks (RNNs) can automatically learn complex and hierarchical feature representations from raw or semi-structured context data \cite{Chen2024Survey}. The Wide \& Deep learning framework by Cheng et al. exemplifies a hybrid approach, combining a wide linear component (often using cross-product features for memorization) with a deep neural network component (using embeddings for generalization) to improve recommendation quality \cite{Cheng2016WideDeep}. Another powerful hybrid technique involves using latent features derived from Matrix Factorization (MF) models as input to contextual bandits. Scores or user/item factors from MF can enrich the context available to the bandit, blending collaborative filtering strengths with the bandit's adaptive learning capabilities \cite{Chen2024ContextualBanditsMovieRec, Tang2014EnsembleContextual, Shariff2018Differentially}. BanditMF explicitly combines MAB with MF \cite{Tang2014EnsembleContextual}, and some systems use clustered user-movie matrices (derived post-SVD) as contextual input for LinUCB \cite{Chen2024ContextualBanditsMovieRec}.

The sophistication of feature engineering in CMAB systems often reflects the complexity of the application domain. While simple, predefined static features may be adequate for controlled simulations or less demanding tasks, real-world, large-scale personalization systems increasingly depend on dynamically learned representations (like embeddings or the outputs of adaptive methods) or hybrid approaches (such as incorporating MF scores) to effectively capture the subtle nuances of user behavior, item characteristics, and their interactions. This progression from handcrafted features to learned representations mirrors broader trends in machine learning, driven by the availability of large datasets and the power of models like deep neural networks to discover salient patterns automatically. Furthermore, there's a growing understanding that for contextual bandits, particularly in personalization, the most valuable features are not merely those correlated with the outcome (e.g., high reward), but those that cause *heterogeneous treatment effects*—that is, features that help differentiate which specific arm (or treatment) is optimal for a given context, rather than just predicting overall success \cite{Hao2024FeatureSelectionCB}. This refined perspective on feature importance is critical for achieving true personalization.

\section{Scope and Assumptions of the Experimental Implementation}

Our experimental implementation applies an online contextual bandit model using logistic regression with stochastic gradient descent (SGD). The reward function models binary outcomes as
\[
P(y=1 \mid x; w) = \sigma(w^T x), \quad \text{where} \quad \sigma(z) = \frac{1}{1 + e^{-z}}.
\]
Here, $x_{t,a}$ denotes the context features for action $a$ at time $t$, which are composed of static, pre-defined user and item characteristics. These may include engineered features such as the Member Purchase Gap (MPG), brand loyalty, and seasonality, as described below:
\begin{align*}
\text{MPG}_c^m &= \frac{d_{\text{event}} - d_{\text{last}}}{T_{cycle}(c, m)} \\
L_{b,c}^m &= \frac{N_{b,c}^m}{\sum_{b'} N_{b',c}^m} \\
S_c(t) &= \frac{f_c(t)}{\max_{t'} f_c(t')}
\end{align*}
where $\text{MPG}_c^m$ is the member purchase gap for category $c$ and member $m$, $L_{b,c}^m$ is the brand loyalty score, and $S_c(t)$ is the seasonality score for category $c$ at time $t$. Additional features include offer recency, duration, nominal discount value, and matrix factorization (MF) scores (used as a bias term to compare multiple offers). Our implementation incorporates both discount value and MF score, but does not explicitly use learned latent representations or dynamic feature adaptation.

For the logistic model, online weight updates are performed using stochastic gradient descent:
\[
w \leftarrow w + \eta (y - \sigma(w^T x)) x
\]
where $\eta$ is the learning rate and $y$ is the observed label. For positive (clipped) samples, the update can be boosted with a multiplier $\alpha > 1$:
\[
w \leftarrow w + \alpha \eta (1 - \sigma(w^T x)) x
\]

The model weights are initialized using domain knowledge to reflect prior beliefs about feature importance when such knowledge is available; otherwise, weights can be set to zero. These initial weights are then backfitted on historical data to calibrate the estimates. This approach combines expert knowledge (when present) with data-driven learning, ensuring the model starts from a reasonable baseline while remaining adaptable to observed patterns.

A key feature of our implementation is that the model weights---and their historical evolution over time---are explicitly tracked and stored for each member. These weight trajectories are not only interpretable by design, but are also made accessible to large language models (LLMs). By surfacing both the current weights and their changes over time, the system enables LLMs to generate detailed, user-level explanations and behavioral personas on demand. LLM can analyze the sequence of weight updates to describe how a member's preferences shift with respect to features like brand loyalty, seasonality, or offer value, providing transparent and actionable insights for stakeholders or end users.

The choice of logistic regression with SGD is motivated not only by its interpretability and theoretical guarantees, but also by its suitability for large-scale, distributed environments. Logistic regression models can be efficiently parallelized and scaled using frameworks like Apache Spark, making them practical for big-data applications where millions of users and offers must be processed in real time. This contrasts with more computationally intensive approaches, such as neural bandits that require deep network training \cite{Zhou2020NeuralUCB}, or Thompson Sampling variants that involve complex posterior inference (e.g., PG-TS for logistic models \cite{Dumitrascu2018PGTS}). While our implementation focuses on this scalable and interpretable baseline, the broader literature explores a range of alternatives, including empirically strong Thompson Sampling methods \cite{ChapelleLi2011EmpiricalTS, AgrawalGoyal2012ThompsonAnalysis} and specialized UCB variants for cold-start or non-stationary settings \cite{Jiang2024HellingerUCB}.

Our implementation's reliance on predefined, static features facilitates a controlled simulation environment. This is distinct from advanced methodologies that focus on adaptive feature extraction, such as the ABaCoDE framework \cite{Lin2018AdaptiveFeature}, or deep learning approaches that automatically discover hierarchical features \cite{Chen2024Survey, Cheng2016WideDeep, Zhou2020NeuralUCB}. We use matrix factorization (MF) scores as a bias term in our logistic regression model, providing a simple way to incorporate collaborative filtering signals; however, we do not learn latent features dynamically as in full hybrid or deep models. Such hybrid approaches are common in production systems \cite{Tang2014EnsembleContextual, Chen2024ContextualBanditsMovieRec}. Temporal dynamics, if addressed, are handled through simple timestamp features or by assuming stationarity, in contrast to models like D-LinUCB that use discounted regression to adapt to drift \cite{Garcelon2020DLinUCB}, or those that explicitly model user purchase rhythms and seasonality \cite{Kooti2016Portrait, Park2024EMKFBandit, Wang2024LinUCBHybrid, Han2021Generalized}.

The assumption of a linear or logistic reward function is a common and useful starting point in CMAB research \cite{Li2010Contextual, AbbasiYadkori2011Improved, ZhangSugiyama2023OnlineMultinomial}, but it can be a simplification of real-world complexities. Much research has focused on relaxing these assumptions through generalized linear models \cite{ZhangSugiyama2023OnlineMultinomial, Filippi2010Parametric} or non-linear function approximators like neural networks \cite{Chen2024Survey, Slivkins2019IntroductionMAB, Zhou2020NeuralUCB}. Our implementation also assumes a stationary environment, where reward distributions and context generation remain fixed, providing a stable setting for analysis but differing from algorithms designed for non-stationary environments \cite{Garcelon2020DLinUCB, Wang2024LinUCBHybrid}.

To support exploration, the system can employ randomized scoring by transforming deterministic probabilities $p_{o,m}$ into sampled scores via Beta distributions:
\[
\tilde{p}_{o,m} \sim \text{Beta}(\alpha, \beta), \quad \alpha = \kappa \cdot p_{o,m}, \quad \beta = \kappa \cdot (1 - p_{o,m})
\]
where $\kappa$ is a scaling factor that can be increased over time to reduce exploration as the campaign progresses.

In summary, our experimental implementation serves as an essential building block or reference point, enabling systematic investigation of core bandit behaviors in a controlled manner. This foundational understanding can be contrasted with more complex, specialized, or empirically-driven solutions found in the literature, which address real-world challenges such as non-stationarity, massive action spaces, nuanced feature interactions, budget constraints \cite{Badanidiyuru2014Resourceful, ZhangHowson2024DISCO}, fairness \cite{Slivkins2019IntroductionMAB}, or complex feedback mechanisms like delayed or sparse rewards and confounded feedback \cite{Zhang2024TSConf}.

Table \ref{tab:script_vs_literature} summarizes the contrast between our implementation's approach and existing research directions.

\begin{table}[h!]
\centering
\caption{Comparison of our approach vs. existing research directions}
\label{tab:script_vs_literature}
\resizebox{\textwidth}{!}{%
\begin{tabular}{p{3.5cm}p{4.5cm}p{5cm}p{3cm}}
\toprule
\textbf{Aspect} & \textbf{Our Approach} & \textbf{Existing Literature} & \textbf{Key Citations} \\
\midrule
Algorithm Type & Logistic Regression with SGD & Neural Bandits, Advanced TS, GLM Bandits (e.g., OFUL-MLogB) & \cite{Zhou2020NeuralUCB, ZhangSugiyama2023OnlineMultinomial, Dumitrascu2018PGTS} \\
Reward Model & Logistic & Non-linear (NNs), Complex GLMs, Non-parametric & \cite{Slivkins2019IntroductionMAB, Zhou2020NeuralUCB, ZhangSugiyama2023OnlineMultinomial, Filippi2010Parametric} \\
Feature Handling & Static, Pre-defined, Basic Encoding & Adaptive Feature Extraction (ABaCoDE), Deep Learned Embeddings, MF Latent Features, Heterogeneous Treatment Effect Features & \cite{Chen2024Survey, Chen2024ContextualBanditsMovieRec, Lin2018AdaptiveFeature, Hao2024FeatureSelectionCB, Cheng2016WideDeep, Tang2014EnsembleContextual} \\
Stationarity & Assumed Stationary & Discounted Regression (D-LinUCB), Change Detection, Models for Temporal Drift/Seasonality & \cite{Garcelon2020DLinUCB, Wang2024LinUCBHybrid, Han2021Generalized} \\
Scale (No. of Arms $K$) & Likely Small to Moderate $K$ & Systems handling thousands/millions of arms & \cite{Zhou2020NeuralUCB} \\
Constraints & Likely Unconstrained & Budgeted Bandits, Resource-Constrained Bandits & \cite{Badanidiyuru2014Resourceful, ZhangHowson2024DISCO} \\
Feedback Complexity & Assumed Direct, Immediate & Delayed Rewards, Sparse Rewards, Confounded/Biased Feedback (Herding) & \cite{Zhang2024TSConf} \\
Other Concerns & Not explicitly modeled & Fairness, Interpretability, Robustness to Adversarial Attacks & \cite{Slivkins2019IntroductionMAB, Zhou2020NeuralUCB} \\
\bottomrule
\end{tabular}%
}
\end{table}

\section{Experiment Design}

The experimental setup implements a contextual bandit approach using logistic regression with stochastic gradient descent (SGD) for online learning. The system processes real retail transaction logs to train and evaluate the model, focusing on predicting clip probabilities for offers based on user and offer features.

\subsection*{Implementation Overview and Methodology}

The system implements a coupon gallery scenario in which, at each round, a user is presented with a set of available offers. Each offer is described by a feature vector that encodes both user- and offer-specific attributes, reflecting the feature engineering described in the whitepaper. These features include the Member Purchase Gap (MPG), brand loyalty score, seasonality score, offer recency and duration, offer value, and a matrix factorization (MF) score. The key innovation in our implementation is the two-level approach:

\begin{itemize}
    \item \textbf{Category Level:} Logistic regression models are trained for each category to predict clip probabilities based on user and offer features. This allows the system to learn category-specific user preferences and behavior patterns.
    \item \textbf{Offer Level:} The category-level predictions are aggregated to the offer level using a weighted combination that takes into account the offer's characteristics and the user's historical behavior.
\end{itemize}

At each round, the bandit algorithm observes the context for all available offers and selects one (or a ranked list) to present. The reward is binary (a clip event), generated according to the ground-truth reward function implemented through the logistic regression models.
All features are normalized using z-score standardization to ensure consistent scale across different feature types. This normalization step is crucial for the logistic regression model to learn effectively from the features.

The implementation evaluates the CAMB algorithm that is first trained on historic (batch) data using scalable systems (e.g., PySpark), and then continues to learn online as new interactions are observed. This mirrors real-world deployment pipelines, where models are periodically retrained offline and then updated incrementally in production. The algorithm uses the engineered features as input, and online updates are performed after each user interaction, following stochastic gradient descent (SGD) for the logistic regression models.

To implement the exploration mechanism, the system uses a stochastic ranking step: after computing deterministic probabilities from the category-level models, a sampled score is drawn from a Beta distribution parameterized by the model's predicted probability and a scaling factor. This introduces controlled randomness into offer selection, especially early in the process, and allows for the study of exploration-exploitation trade-offs.
The implementation processes real retail transaction logs, replaying historical data to train and evaluate the model. This approach allows us to study the model's performance on actual user behavior patterns while maintaining the ability to analyze learning dynamics in a controlled manner.

\subsection*{Implementation Procedure and Evaluation}

The implementation procedure follows a systematic approach to evaluate the model's performance. The process involves loading and processing real transaction logs, running multiple rounds where the model selects offers based on contextual features, updating the model online with observed feedback, and logging key metrics for analysis.

The overall implementation workflow is summarized in the following algorithm. It outlines the key steps of the process, including initialization of users and offers, iterative selection and feedback by the contextual bandit algorithm, and online model updates. This step-by-step process enables systematic evaluation of learning dynamics, exploration-exploitation trade-offs, and model adaptation over time.

\begin{figure}[h]
\centering
\begin{minipage}{0.95\linewidth}
\begin{algorithm}[H]
\caption{Implementation and Evaluation Procedure for proposed CAMB System}
\begin{algorithmic}[1]
\State \textbf{Initialization:} Load real retail transaction logs and initialize feature scaler.
\For{each round $t = 1, \ldots, T$}
    \State Process a user and their available offers from historical data to simulate online events.
    \State Compute and normalize feature vectors for each offer.
    \State Predict clip probabilities using logistic regression.
    \State Select offers based on predicted probabilities.
    \State Process the clip event using the reward function.
    \State Update the category-level logistic regression models.
    \State Detect any significant changes in model coefficients and generate insights with LLM by analyzing the updated model weights.
    \State Log predictions, rewards, and model weights.
\EndFor
\State \textbf{Analysis:} Aggregate results over multiple runs to report average cumulative reward, regret, and feature weight dynamics.
\end{algorithmic}
\end{algorithm}
\end{minipage}
\end{figure}

\begin{figure}[h]
    \centering
    \includegraphics[width=\textwidth]{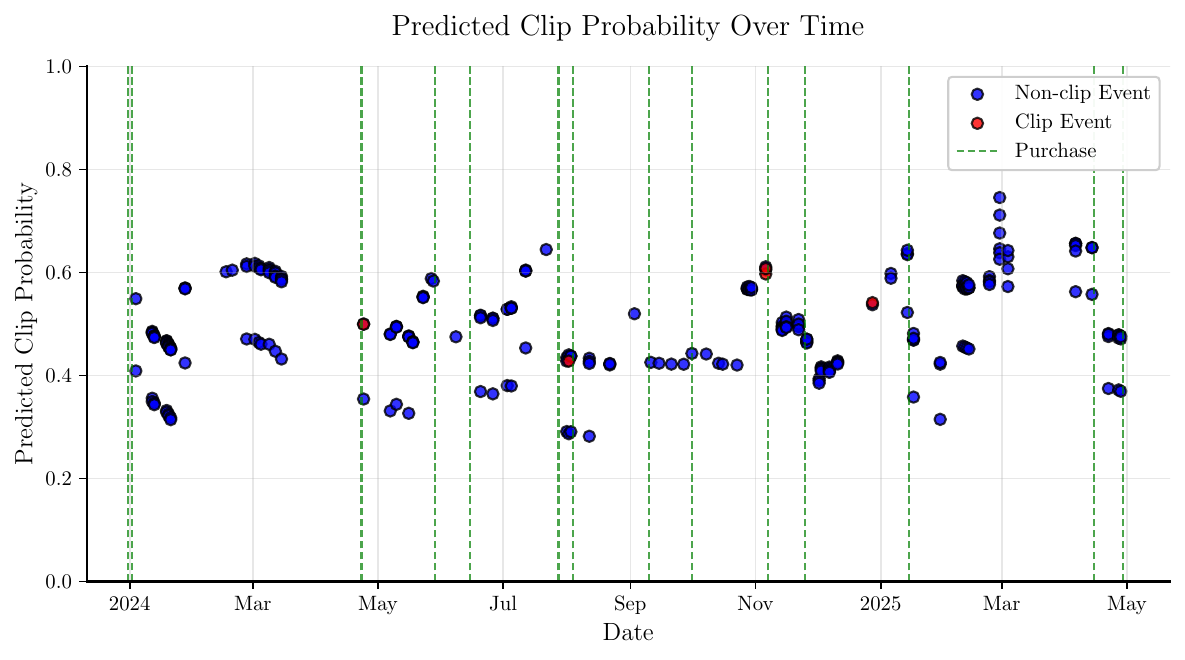}
    \caption{Predicted clip probabilities for a single category present in all offers over time, showing both clip (red) and non-clip (blue) events, with purchase dates marked by green vertical lines.}
    \label{fig:clip_prob}
\end{figure}

Figure \ref{fig:clip_prob} shows the predicted clip probabilities for a single category present in all offers over time, with blue dots representing non-clip events and red dots representing clip events. The green vertical lines indicate purchase dates. This visualization helps us understand how well the model predicts user behavior and how these predictions evolve over time.

\begin{figure}[h]
    \centering
    \includegraphics[width=\textwidth]{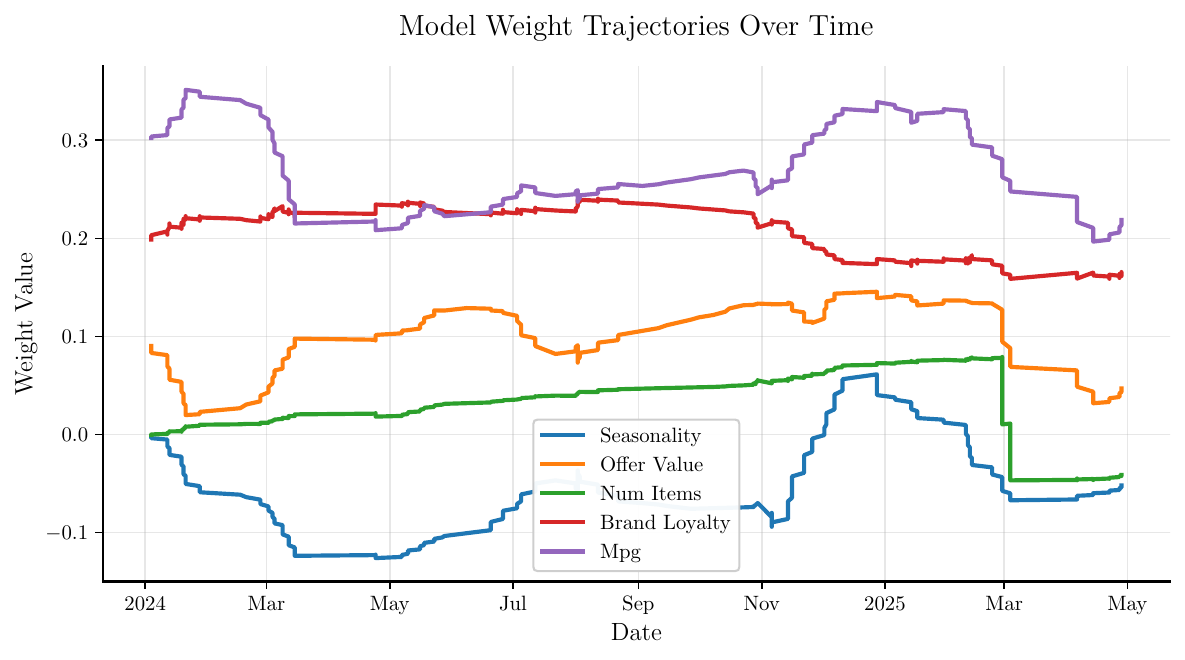}
    \caption{Model weight trajectories over time for an example user-category pair, showing how different features' influence evolves during the simulation.}
    \label{fig:weight_traj}
\end{figure}

Figure \ref{fig:weight_traj} displays the evolution of model weights over time for an example user-category pair, providing insights into how different features influence the model's predictions. The trajectories show how the model learns and adapts to user behavior patterns.

A key advantage of our interpretable approach is the ability to generate detailed member profiles using AI to analyze these weight trajectories. Our large language model can analyze the weight patterns shown in Figure \ref{fig:weight_traj} to produce insights such as:

\begin{quote}
\small\itshape
This member's behavior in the category is primarily driven by the time since their last purchase, as indicated by the high and fluctuating weight on the MPG (member purchase gap) feature. They tend to clip offers when a longer gap has passed, suggesting clip behavior is replenishment-driven. Brand loyalty also plays a strong and stable role, indicating a preference for familiar brands, especially when the purchase gap is not large.

Offer value has a moderate but consistent influence, meaning that while deep discounts are not essential, they can help sway decisions when timing or brand signals are less clear. Seasonality generally has a negative weight, implying that the member's activity in this category does not align with typical seasonal patterns. Instead, their behavior is more idiosyncratic or based on personal needs. Over time, the number of items in an offer has become more relevant, possibly reflecting a shift toward buying in bulk or selecting offers that cover more needs in a single clip.

Overall, this is a brand-loyal, non-seasonal member whose clip behavior is timed around replenishment cycles and increasingly influenced by offer size rather than pure discount depth.
\end{quote}

Such AI-generated interpretations enable personalized offer optimization by identifying key behavioral drivers for each member, going beyond traditional metrics to provide actionable insights for marketing strategy.

Performance is evaluated using several key metrics: cumulative reward (the total number of successful offer redemptions over all rounds and users), regret (the difference between the cumulative reward obtained by the algorithm and that of an oracle policy that always selects the offer with the highest true reward probability), optimal action rate (the proportion of rounds in which the algorithm selected the offer with the highest ground-truth probability), model coefficient trajectories (tracking the evolution of model weights for interpretable analysis), and per-round reward (the average reward per round, visualized over time to assess learning progress and convergence).

\subsection*{Application to Large-Scale System}

This experimental design demonstrates the core components of the proposed solution in a controlled, small-scale environment. While the current implementation processes a subset of real retail transaction logs, the approach is designed to scale to production-level deployment. The system uses the same behavioral and offer features (MPG, brand loyalty, seasonality, offer value, MF score) and learning mechanisms (online SGD updates, stochastic sampling) that would be used in a full-scale implementation.

The current implementation serves as a proof-of-concept, validating the key design decisions and learning dynamics of the approach. While we track both operational metrics (such as clip probability and ranking accuracy) and model interpretability (such as weight traces), a full-scale deployment would require additional considerations:

\begin{itemize}
    \item Distributed training and serving infrastructure for handling millions of users and offers
    \item Real-time feature computation and model updates at scale
    \item A/B testing framework for controlled rollout and evaluation
    \item Monitoring and alerting systems for model performance and data quality
\end{itemize}

Future work will focus on scaling this approach to production, including:
\begin{itemize}
    \item Implementing distributed training using PySpark or similar frameworks
    \item Developing a real-time serving infrastructure for low-latency predictions
    \item Conducting large-scale A/B tests to validate the approach with real users
    \item Building comprehensive monitoring and evaluation pipelines
\end{itemize}

The current small-scale implementation provides a clear, reproducible baseline for understanding the behavior of the approach before deploying to a full-scale production environment. The results demonstrate the potential of the method, but final validation of its effectiveness will require large-scale deployment and evaluation.

\section{Limitations and Future Work}

This proposal presents an early-stage prototype for interpretable contextual bandits in multi-category retail environments. While we believe the framework offers a promising balance between transparency and modular learning, several limitations should be acknowledged to guide future work.

At this stage, the system has not yet been validated under production-scale conditions. Key performance characteristics—such as latency, throughput, memory usage, and convergence behavior—remain to be evaluated in high-load environments. Although the implementation has been designed with scalability in mind, future testing in distributed infrastructures such as Apache Spark or Flink, as well as in real-time serving contexts, will be necessary to confirm its operational viability.

Additionally, the predictive and reward optimization performance of the approach has not been benchmarked against established baselines. Comparative analysis with methods such as Logistic Thompson Sampling or LinUCB, and the use of offline evaluation techniques with logged bandit feedback, would allow for a more thorough understanding of the trade-offs between interpretability and effectiveness. We also note that formal regret analysis has not yet been conducted and remains an important area for theoretical development.

The current feature set is grounded in domain-relevant constructs such as Member Purchase Gap (MPG) and Matrix Factorization (MF) scores, yet their empirical utility under real-world distribution shifts requires further validation. Our assumption of relatively stable user preferences and market conditions simplifies the learning problem but may limit applicability in more dynamic settings. Incorporating temporal adaptation mechanisms—such as sliding windows, discounting, or variants of D-LinUCB—will be a key focus moving forward.

The exploration strategy, while intuitive and easy to implement, currently relies on heuristics without formal guarantees. Exploring more theoretically grounded alternatives, including confidence-based methods or bootstrapped uncertainty estimates, may improve both learning efficiency and robustness.

Finally, although interpretability is central to the framework, we have not yet formally evaluated how model outputs—such as weight trajectories—are perceived by end users or decision-makers. Future work will benefit from qualitative assessments or user studies to better understand the role of transparency in practice.

Looking ahead, we see several important directions for extending this work. Empirical validation using A/B tests and offline replay will be crucial for assessing effectiveness. Scaling the system to support larger user bases, offer catalogs, and feature spaces will require performance-focused engineering. We also plan to explore richer feature representations, including deep embeddings and hybrid models that maintain interpretability while increasing flexibility. Addressing non-stationarity and supporting multi-objective reward functions, such as long-term customer value or category-level optimization, are also important avenues for development.

In summary, while the current framework establishes a transparent and theoretically motivated foundation for contextual bandits in retail, we recognize that substantial empirical and technical refinement is necessary. We hope that this work serves as a solid starting point for more robust, scalable, and trustworthy decision systems in real-world retail applications.

\section{Conclusion}
The field of contextual bandits is dynamic and rapidly advancing, with research continually expanding the frontiers of algorithmic sophistication, theoretical insight, and real-world applicability to complex personalization challenges. Foundational algorithms such as LinUCB, Epsilon-Greedy, and Thompson Sampling, together with reward models like linear and logistic regression, provide the essential underpinnings of this area. Recent work increasingly emphasizes scalability—enabling solutions to operate efficiently at the scale of millions of users and offers—and the interpretability of models, particularly through the use of large language models (LLMs) to surface actionable insights from model parameters. Our experimental implementation demonstrates a scalable and interpretable contextual bandit framework: it leverages logistic regression with static features in a stationary environment, and crucially, exposes model weights in a form that can be directly interpreted by LLMs for transparent, user-level explanations. This approach offers a controlled environment for investigating bandit behavior, while also highlighting the potential for interpretable, AI-powered personalization at production scale, thus situating our findings within the broader and evolving landscape of contextual bandit research.

\section*{Acknowledgments}

This research is (partly) supported by "European Digital Innovation Hub Adriatic Croatia (EDIH Adria) (project no. 101083838)" under the European Commission's Digital Europe Programme, SPIN project "INFOBIP Konverzacijski Order Management (IP.1.1.03.0120)", SPIN project “Projektiranje i razvoj nove generacije laboratorijskog informacijskog sustava (iLIS)" (IP.1.1.03.0158), SPIN project “Istraživanje i razvoj inovativnog sustava preporuka za napredno gostoprimstvo u turizmu (InnovateStay)" (IP.1.1.03.0039), and the FIPU project "Sustav za modeliranje i provedbu poslovnih procesa u heterogenom i decentraliziranom računalnom sustavu”.

The authors acknowledge the use of artificial intelligence tools for text styling and formatting assistance. However, the authors take full responsibility for the content, accuracy, and scientific integrity of this work. 

\clearpage

\bibliography{references.bib}
\bibliographystyle{unsrt}

\end{document}